\documentclass{article}
\usepackage[T1]{fontenc} 
\usepackage[utf8]{inputenc} 
\usepackage{ismir,amsmath,cite,url}
\usepackage{graphicx}
\usepackage{color}
\usepackage[dvipsnames]{xcolor}
\usepackage{units}
\usepackage{gensymb}
\usepackage{paralist}
\usepackage{amssymb}
\usepackage{amsfonts}
\usepackage{multirow}
\usepackage{booktabs}
\usepackage{tabularx}
\usepackage{subcaption}
\usepackage{multicol}
\usepackage{microtype}
\usepackage{flushend}

\newcommand{\ashis}[1]{#1}

\title{\MakeLowercase{d}Melodies: A Music Dataset for Disentanglement Learning}

\multauthor
{Ashis Pati \hspace{1cm} Siddharth Gururani \hspace{1cm} Alexander Lerch} 
{
  Center for Music Technology, Georgia Institute of Technology, USA\\
  {\tt\small ashis.pati@gatech.edu, siddgururani@gatech.edu, alexander.lerch@gatech.edu}
}
\def\authorname{Ashis Pati, Siddharth Gururani, Alexander Lerch}

\begin{document}
\maketitle
\begin{abstract}
  Representation learning focused on disentangling the underlying factors of variation in given data has become an important area of research in machine learning. However, most of the studies in this area have relied on datasets from the computer vision domain and thus, have not been readily extended to music. In this paper, we present a new symbolic music dataset that will help researchers working on disentanglement problems demonstrate the efficacy of their algorithms on diverse domains. This will also provide a means for evaluating algorithms specifically designed for music. To this end, we create a dataset comprising of 2-bar monophonic melodies where each melody is the result of a unique combination of nine latent factors that span ordinal, categorical, and binary types. The dataset is large enough ($\approx$ 1.3 million data points) to train and test deep networks for disentanglement learning. In addition, we present benchmarking experiments using popular unsupervised disentanglement algorithms on this dataset and compare the results with those obtained on an image-based dataset. 
\end{abstract}

\section{Introduction}
\label{sec:intro}
\begin{figure}[t]
    \centering
    \includegraphics[width =1\columnwidth]{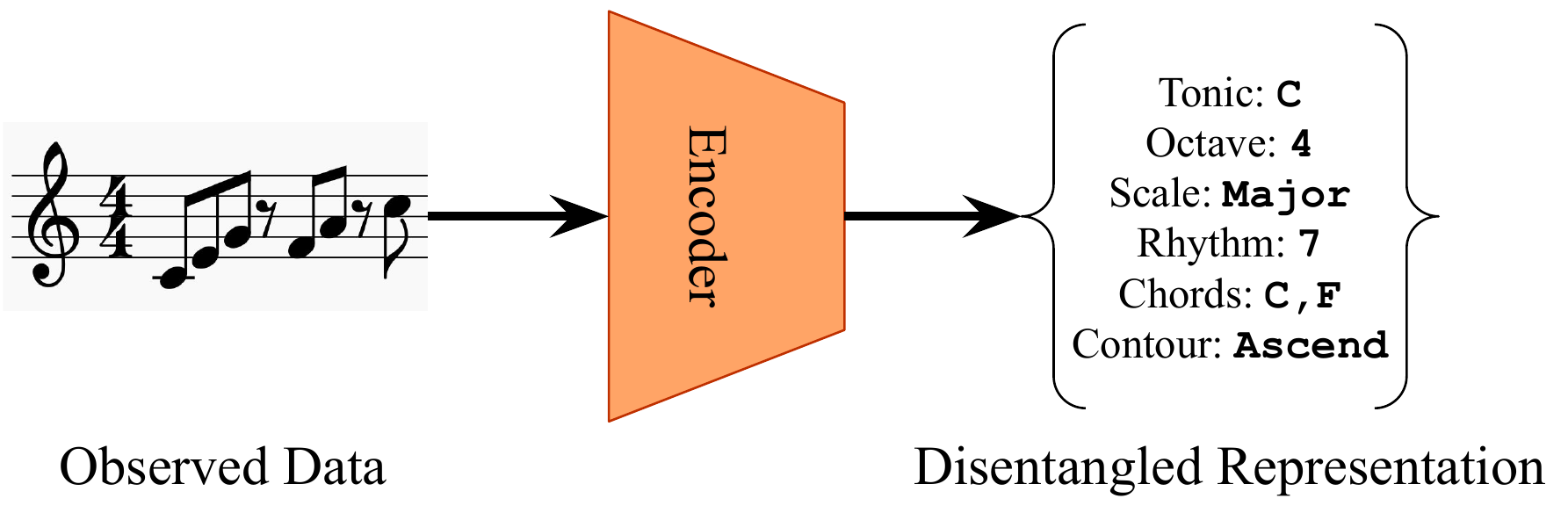}
    \caption{Disentanglement example where a high dimensional observed data is disentangled into a low dimensional representation comprising of semantically meaningful factors of variation. 
    }
    \label{fig:disent_example}
  \end{figure}

  Representation learning deals with extracting the underlying factors of variation in a given observation \cite{bengio_representation_2013}. Learning compact and \textit{disentangled} representations (see ~\figref{fig:disent_example} for an illustration) from given data, where important factors of variation are clearly separated, is considered useful for generative modeling and for improving performance on downstream tasks (such as speech recognition, speech synthesis, vision and language generation \cite{hsu2017unsupervised,hsu2019disentangling,kexin2018neural}). Disentangled representations allow a greater degree of interpretability and controllability, especially for content generation, be it language, speech, or music. In the context of Music Information Retrieval (MIR) and generative music models, learning some form of disentangled representation has been the central idea for a wide variety of tasks such as genre transfer \cite{brunner_midi-vae_2018}, rhythm transfer \cite{yang2019deep,jiang2020transformer}, timbre synthesis \cite{luo2019learning}, instrument rearrangement \cite{hung2019musical}, manipulating musical attributes \cite{hadjeres_glsr-vae_2017,pati19latent-reg}, and learning music similarity \cite{lee2020disentangled}. 

  Consequently, there exists a large body of research in the machine learning community focused on developing algorithms for learning disentangled representations. These span unsupervised \cite{higgins_beta-vae_2017,chen_isolating_2018,kim_disentangling_2018,kumar_variational_2017}, semi-supervised \cite{kingma2014semi, siddharth2017learning, Locatello2020Disentangling} and supervised \cite{lample_fader_2017,hadjeres_glsr-vae_2017,kulkarni_deep_2015,donahue_semantically_2018} methods. However, a vast majority of these algorithms are designed, developed, tested, and evaluated using data from the image or computer vision domain. 
  The availability of standard image-based datasets such as dSprites~\cite{matthey_dsprites_2017}, 3D-Shapes~\cite{burges_3d-shapes_2020}, and 3D-Chairs~\cite{aubry_seeing_2014} among others has fostered disentanglement studies in vision.
  Additionally, having well-defined factors of variation (for instance, size and orientation in dSprites \cite{matthey_dsprites_2017}, pitch and elevation in Cars3D \cite{reed_deep_2015}) has allowed systematic studies and easy comparison of different algorithms. 
  However, this restricted focus on a single domain raises concerns about the generalization of these methods \cite{locatello_challenging_2019} and prevents easy adoption into other domains such as music. 
  
  Research on disentanglement learning in music has often been application-oriented with researchers using their own problem-specific datasets. The factors of variation have also been chosen accordingly. To the best of our knowledge, there is no standard dataset for disentanglement learning in music. This has prevented systematic research on understanding disentanglement in the context of music. 
  
  In this paper, we introduce \textit{dMelodies}, a new dataset of monophonic melodies, specifically intended for disentanglement studies. The dataset is created algorithmically and is based on a simple and yet diverse set of independent latent factors spanning ordinal, categorical and binary attributes. The full dataset contains $\approx 1.3$ million data points which matches the scale of image datasets and should be sufficient to train deep networks. We consider this dataset as the primary contribution of this paper. In addition, we also conduct benchmarking experiments using three popular unsupervised methods for disentanglement learning and present a comparison of the results with the dSprites dataset \cite{matthey_dsprites_2017}. Our experiments show that disentanglement learning methods do not directly translate between the image and music domains and having a music-focused dataset will be extremely useful to ascertain the generalizability of such methods. The dataset is available online\footnote{https://github.com/ashispati/dmelodies\_dataset} along with the code to reproduce our benchmarking experiments.\footnote{\ashis{https://github.com/ashispati/dmelodies\_benchmarking}}
  

\section{Motivation}
\label{sec:motivation}
  In representation learning, given an observation $\mathbf{x}$, the task is to learn a representation $r(\mathbf{x})$ which ``makes it easier to extract useful information when building classifiers or other predictors'' \cite{bengio_representation_2013}. The fundamental assumption is that any high-dimensional observation $\mathbf{x} \in \mathcal{X}$ (where $\mathcal{X}$ is the data-space) can be decomposed into a semantically meaningful low dimensional latent variable $\mathbf{z} \in \mathcal{Z}$ (where $\mathcal{Z}$ is referred to as the latent space). Given a large number of observations in $\mathcal{X}$, the task of disentanglement learning is to estimate this low dimensional latent space $\mathcal{Z}$ by separating out the distinct factors of variation \cite{bengio_representation_2013}. An ideal disentanglement method ensures that changes to a single underlying factor of variation in the data changes only a single factor in its representation \cite{locatello_challenging_2019}. From a generative modeling perspective, it is also important to learn the mapping from $\mathcal{Z}$ to $\mathcal{X}$ to enable better control over the generative process. 
  
  \subsection{Lack of diversity in disentanglement learning}
    Most state-of-the-art methods for unsupervised disentanglement learning are based on the Variational Auto-Encoder (VAE) \cite{kingma_auto-encoding_2014} framework. The key idea behind these methods is that factorizing the latent representation to have an aggregated posterior should lead to better disentanglement \cite{locatello_challenging_2019}. This is achieved using different means, e.g., imposing constraints on the information capacity of the latent space \cite{higgins_beta-vae_2017,burgess_understanding_2018,rubenstein_learning_2018}, maximizing the mutual information between a subset of the latent code and the observations \cite{chen_infogan_2016}, and maximizing the independence between the latent variables \cite{chen_isolating_2018,kim_disentangling_2018}. However, unsupervised methods for disentanglement learning are sensitive to inductive biases (such network architectures, hyperparameters, and random seeds) and consequently there is a need to properly evaluate such methods by using datasets from diverse domains \cite{locatello_challenging_2019}.
    
    
    Apart from unsupervised methods for disentanglement learning, there has also been some research on semi-supervised~\cite{siddharth2017learning, Locatello2020Disentangling} and supervised~\cite{kulkarni_deep_2015,lample_fader_2017,connor2019representing,engel_latent_2017} learning techniques to manipulate specific attributes in the context of generative models. In these paradigms, a labeled loss is used in addition to the unsupervised loss. Available labels can be utilized in various ways. 
    They can help with disentangling known factors (e.g., digit class in MNIST) from latent factors (e.g., handwriting style) \cite{bouchacourt_multi-level_2018}, or supervising specific latent dimensions to map to specific attributes \cite{hadjeres_glsr-vae_2017}.
    However, most of these approaches are evaluated using image domain datasets.

    Tremendous interest from the machine learning community has led to the creation of benchmarking datasets (albeit image-based) specifically targeted towards disentanglement learning such as dSprites \cite{matthey_dsprites_2017}, 3D-Shapes \cite{burges_3d-shapes_2020}, 3D-chairs \cite{aubry_seeing_2014}, MPI3D \cite{gondal2019transfer}, most of which are artificially generated and have simple factors of variation. While one can argue that artificial datasets do not reflect real-world scenarios, the relative simplicity of these datasets is often desirable since they enable rapid prototyping. 

  \subsection{Lack of consistency in music-based studies}
  Representation learning has also been explored in the field of MIR. Much like images, learning better representations has been shown to work well for MIR tasks such as composer classification \cite{bretan15learning,gururani2019comparison}, music tagging \cite{choi2017transfer}, and audio-to-score alignment \cite{lattner2019learning}. The idea of disentanglement has been particularly gaining traction in the context of interactive music generation models \cite{engel_latent_2017,brunner_midi-vae_2018,yang2019deep,pati19latent-reg}. Disentangling semantically meaningful factors can significantly improve the usefulness of music generation tools. Many researchers have independently tried to tackle the problem of disentanglement in the context of symbolic music by using different musically meaningful attributes such as genre \cite{brunner_midi-vae_2018}, note density \cite{hadjeres_glsr-vae_2017}, rhythm \cite{yang2019deep}, and timbre \cite{luo2019learning}. However, these methods and techniques have all been evaluated using different datasets which makes a direct comparison impossible. Part of the reason behind this lack of consistency is the difference in the problems that these methods were looking to address. However, the availability of a common dataset allowing researchers to easily compare algorithms and test their hypotheses will surely aid systematic research.

     

\section{\MakeLowercase{d}Melodies Dataset}
\label{sec:design}
  The primary objective of this work is to create a simple dataset for music disentanglement that can alleviate some of the shortcomings mentioned in \secref{sec:motivation}: first, researchers interested in disentanglement will have access to more diverse data to evaluate their methods, and second, research on music disentanglement will have the means for conducting systematic, comparable evaluation. This section describes the design choices and the methodology used for creating the proposed \textit{dMelodies} dataset. 

  While core MIR tasks such as music transcription, or tagging focus more on analysis of audio signals, research on generative models for music has focused more on the symbolic domain. Considering most of the interest in disentanglement learning stems from research on generative models, we decided to create this dataset using symbolic music representations. 

  \subsection{Design Principles}
  \label{sec:design_principles}
    To enable objective evaluation of disentanglement algorithms, one needs to either know the ground-truth values of the underlying factors of variation for each data point, or be able to synthesize the data points based on the attribute values. The dSprites dataset \cite{matthey_dsprites_2017}, for instance, consists of single images of different 2-dimensional shapes with simple attributes specifying the position, scale and orientation of these shapes against a black background. The design of our dataset is loosely based on the dSprites dataset. The following principles were used to finalize other design choices:
    \begin{compactenum}[(a)]
      \item The dataset should have a simple construction with homogenous data points and intuitive factors of variation. It should allow for easy differentiation between data points and have clearly distinguishable latent factors.
      \item The factors of variation should be independent, i.e., changing any one factor should not cause changes to other factors. While this is not always true for real-world data, it enables consistent objective evaluation. 
      \item There should be a clear one-to-one mapping between the latent factors and the individual data points. In other words, each unique combination of the factors should result in a unique data point. 
      \item The factors of variation should be diverse. In addition, it would be ideal to have the factors span different types such as discrete, ordinal, categorical and binary.
      \item Finally, the different combinations of factors should result in a dataset large enough to train deep neural networks. Based on size of the different image-based datasets \cite{matthey_dsprites_2017,liu_deep_2015}, we would require a dataset of the order of at least a few hundred thousand data points.
    \end{compactenum}

  \subsection{Dataset Construction}
    Considering the design principles outlined above, we decided to focus on monophonic pitch sequences. While there are other options such as polyphonic or multi-instrumental music, the choice of monophonic melodies was to ensure simplicity. Monophonic melodies are a simple form of music uniquely defined by the pitch and duration of their note sequences. The pitches are typically based on the key or scale in which the melody is being played and the rhythm is defined by the onset positions of the notes.
    
    Since the set of all possible monophonic melodies is very large and heterogeneous, the following additional constraints were imposed on the melody in order to enforce homogeneity and satisfy the other design principles:
    \begin{compactenum} [(a)]
      \item Each melody is based on a scale selected from a finite set of allowed scales. This choice of scale also serves as one of the factors of variation. The melody will also be uniquely defined by the pitch class of the tonic (root pitch) and the octave number.
      \item In order to constrain the space of all possible pitch patterns within a scale, we restrict each melody to be an arpeggio over the standard I-IV-V-I cadence chord pattern. Consequently, each melody consists of 12 notes (3 notes for each of the 4 chords).
      \item In order to vary the pitch patterns, the direction of arpeggiation of each chord, i.e. up or down, is used as a latent factor. This choice adds a few binary factors of variation to the dataset. 
      \item The melodies are fixed to 2-bar sequences with 8th note as the minimum note duration. This makes the dataset uniform in terms of sequence lengths of the data points and also helps reduce the complexity of the sequences. 2-bar sequences have been used in other music generation studies as well \cite{hadjeres_glsr-vae_2017,roberts_hierarchical_2018}. 
      We use a tokenized data representation such that each melody is a sequence of length 16. 
      \item If we consider the space of all possible unique rhythms, the number of options will explode to $16 \choose 12$ which will be significantly larger than other factors of variation. Hence, we choose to break the latent factor for rhythm into 2 independent factors: rhythm for bar 1 and bar 2. 
      \item The rhythm of a melody is based on the metrical onset position of the notes \cite{toussaint_mathematical_2002}. Consequently, rhythm is dependent on the number of notes. In order to keep rhythm independent from other factors, we constrain each bar to have 6 notes (play 2 chords) thereby obtaining $8 \choose 6$ options for each bar.  
    \end{compactenum}

\begin{figure}[t]
      \centering
      \includegraphics[width =1\columnwidth]{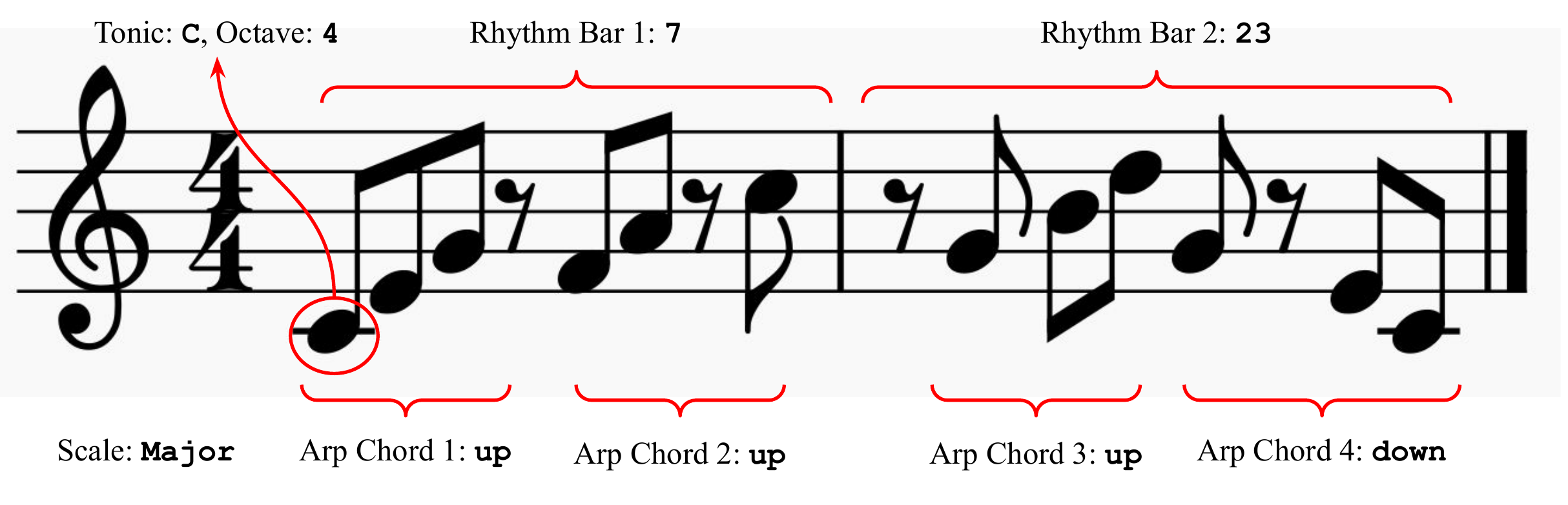}
      \caption{Example of a sample melody from the dMelodies dataset. Also shown are the values of the different latent factors. For rhythm latent factors, the shown value corresponds to the index from the rhythm dictionary.}
      \label{fig:dataset_example}
    \end{figure}

    Based on the above design choices, the dMelodies dataset consists of 2-bar monophonic melodies with 9 factors of variations listed in \tabref{tab:factor_of_variation}. 
    \ashis{The factors of variation were chosen to satisfy the design principles listed in \secref{sec:design_principles}. For instance, while melodic transformations such as repetition, inversion, retrograde would have made more musical sense, they did not allow creation of a large-enough dataset with independent factors of variation. 
    The resulting dataset thus contains simple melodies which do not adequately reflect real-world musical data. A side-effect of this choice of factors is that some of them (such as arpeggiation direction and rhythm) affect only a specific part of the data.} Since each unique combination of these factors results in a unique data point we get 1,354,752 unique melodies. \figref{fig:dataset_example} shows one such melody from the dataset and its corresponding latent factors. \ashis{The dataset is generated using the \textit{music21} \cite{cuthbert_music21_2010} python package.} 

    \begin{table}[t]
      \footnotesize
      \begin{center}
      \begin{tabularx}{\columnwidth}{lcl}
          \toprule
          \textbf{Factor}       & \textbf{$\#$ Options} & \textbf{Notes} \\
          \toprule
          \textit{Tonic}        & 12 & C, C$\#$, D, through B   \\ \midrule
          \textit{Octave}       & 3 & Octave 4, 5 and 6  \\ \midrule
          \textit{Scale}        & 3 & major, harmonic minor, and blues   \\ \midrule
          \textit{Rhythm Bar 1} & 28 & $8 \choose 6$, based on onset locations of 6 notes  \\ \midrule
          \textit{Rhythm Bar 2} & 28 & $8 \choose 6$, based on onset locations of 6 notes  \\ \midrule
          \textit{Arp Chord 1}  & 2 & up/down, for Chord 1 \\ \midrule
          \textit{Arp Chord 2}  & 2 & up/down, for Chord 2 \\ \midrule
          \textit{Arp Chord 3}  & 2 & up/down, for Chord 3 \\ \midrule
          \textit{Arp Chord 4}  & 2 & up/down, for Chord 4 \\ \bottomrule
      \end{tabularx}
      \end{center}
    \caption{Table showing the different factors of variation for the dMelodies dataset. Since all factors of variation are independent, the total dataset contains 1,354,752 unique melodies.}
    \label{tab:factor_of_variation}
    \end{table} 

\section{Benchmarking Experiments}
\label{sec:exp}
  In this section, we present benchmarking experiments to demonstrate the performance of some of the existing unsupervised disentanglement algorithms on the proposed dMelodies dataset and contrast the results with those obtained on the image-based dSprites dataset. 
  
  \subsection{Experimental Setup}
    We consider 3 different disentanglement learning methods: $\beta$-VAE \cite{higgins_beta-vae_2017}, Annealed-VAE \cite{burgess_understanding_2018}, and FactorVAE \cite{kim_disentangling_2018}. All these methods are based on different regularization terms applied to the VAE loss function. 

    \subsubsection{Data Representation}
      We use a tokenized data representation \cite{hadjeres2017deepbach} with the 8th-note as the smallest note duration. Each 8th note position is encoded with a token corresponding to the note name which starts on that position. A special continuation symbol (`\_\_') is used which denotes that the previous note is held. A special token is used for rest. 
  
    \subsubsection{Model Architectures} 
      Two different VAE architectures are chosen to conduct these experiments. The first architecture (dMelodies-CNN) is based on Convolutional Neural Networks (CNNs) and is similar to those used for several image-based VAEs, except that we use 1-D convolutions. The second architecture (dMelodies-RNN) is based on a hierarchical recurrent model \cite{roberts_hierarchical_2018,pati_learning_2019}. 
      Details of the model architectures are provided in the supplementary material. 

    \subsubsection{Hyperparameters}
      Each learning method has its own regularizing hyperparameter. For $\beta$-VAE, we use three different values of $\beta \in \left\{ 0.2, 1.0, 4.0 \right\}$. This choice is loosely based on the notion of normalized-$\beta$ \cite{higgins_beta-vae_2017}. In addition, we force the KL-regularization only when the KL-divergence exceeds a fixed threshold $\tau=50$ \cite{kingma_improved_2016,roberts_hierarchical_2018}. For Annealed-VAE, we fix $\gamma=1.0$ and use three different values of capacity, $C \in \left\{ 25.0, 50.0, 75.0 \right\}$. For FactorVAE, we use the Annealed-VAE loss function with a fixed capacity ($C = 50$), and choose three different values for $\gamma \in \left\{ 1, 10, 50 \right\}$.

    \subsubsection{Training Specifications}
      For each of the above methods, model, and hyperparameter combination, we train 3 models with different random seeds. To ensure consistency across training, all models are trained with a batch-size of $512$ for $100$ epochs. The ADAM optimizer \cite{kingma_adam_2015} is used with a fixed learning rate of $1\mbox{e\ensuremath-}4$, $\beta _{1}=0.9$, $\beta _{2}=0.999$, and $\epsilon = 1\mbox{e\ensuremath-}8$. For $\beta$-VAE and Annealed-VAE, we use 10 warm-up epochs where $\beta=0.0$. After warm-up, the regularization hyperparameter ($\beta$ for $\beta$-VAE and $C$ for Annealed-VAE) is annealed exponentially from $0.0$ to their target values over $100000$ iterations. For FactorVAE, we stick to the original implementation and do not anneal any of the parameters in the loss function. The VAE optimizer is the same as mentioned earlier. The FactorVAE discriminator is optimized using ADAM with a fixed learning rate of $1\mbox{e\ensuremath-}4$, $\beta _{1}=0.8$, $\beta _{2}=0.9$, and $\epsilon = 1\mbox{e\ensuremath-}8$. We found that utilizing the original hyperparameters \cite{kim_disentangling_2018} for this optimizer led to unstable training on dMelodies.

      For comparison with dSprites, we present the results for all the three methods using a CNN-based VAE architecture. The set of hyperparameters and other training configurations were kept the same for the dSprites dataset, except for the FactorVAE where we use the originally proposed loss function and discriminator optimizer hyperparameters, as the model does not converge otherwise.
  
    \subsubsection{Disentanglement Metrics}
      The following objective metrics for measuring disentanglement are used:
      \begin{inparaenum}[(a)]
        \item \textit{Mutual Information Gap (MIG)} \cite{chen_isolating_2018}, which measures the difference of mutual information between a given latent factor and the top two dimensions of the latent space which share maximum mutual information with the factor,
        \item \textit{Modularity} \cite{ridgeway_learning_2018}, which measures if each dimension of the latent space depends on only one latent factor, and 
        \item \textit{Separated Attribute Predictability (SAP)} \cite{kumar_variational_2017}, which measures the difference in the prediction error of the two most predictive dimensions of the latent space for a given factor.
      \end{inparaenum}
      For each metric, the mean across all latent factors is used for aggregation. For consistency, standard implementations of the different metrics are used \cite{locatello_challenging_2019}.

  \subsection{Experimental Results}

    \subsubsection{Disentanglement}

      \begin{figure*}[t]
        \centering
        \begin{tabular}{@{}c@{}}
          \includegraphics[width=0.33\textwidth]{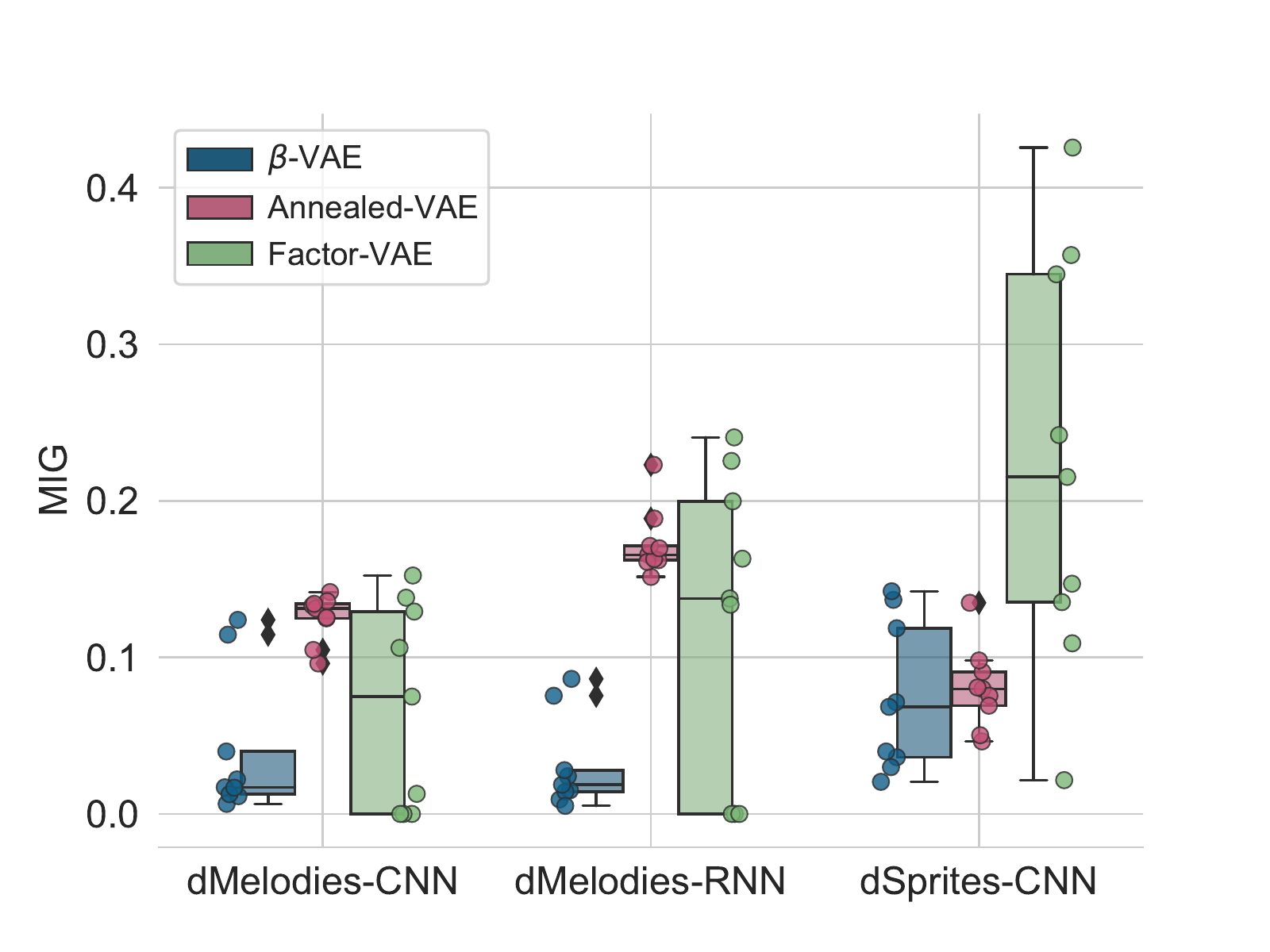} \\[\abovecaptionskip]
          \small (a) \textit{MIG}
        \end{tabular}
        \begin{tabular}{@{}c@{}}
          \includegraphics[width=0.33\textwidth]{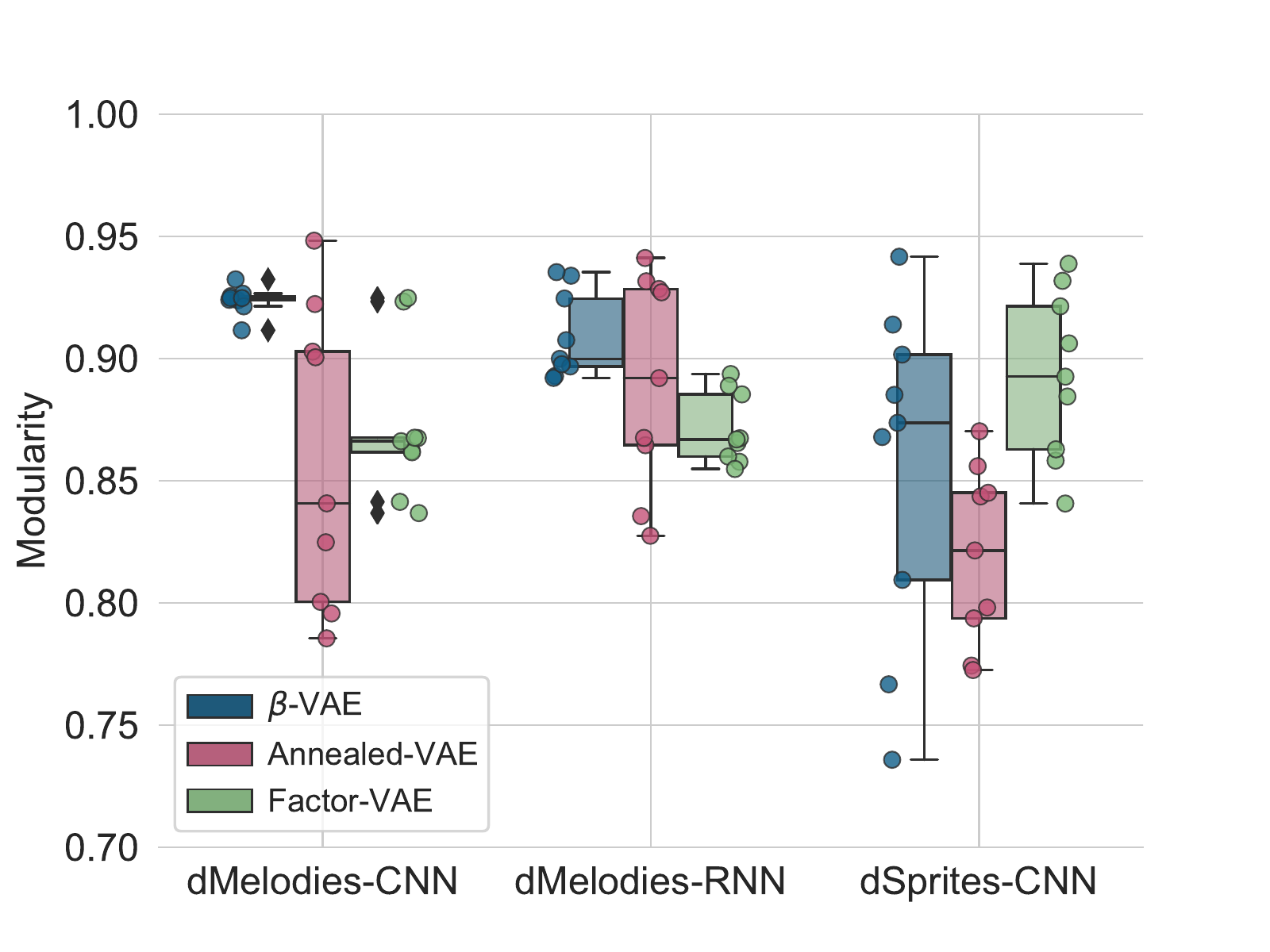} \\[\abovecaptionskip]
          \small (b) \textit{Modularity}
        \end{tabular}
        \begin{tabular}{@{}c@{}}
          \includegraphics[width=0.33\textwidth]{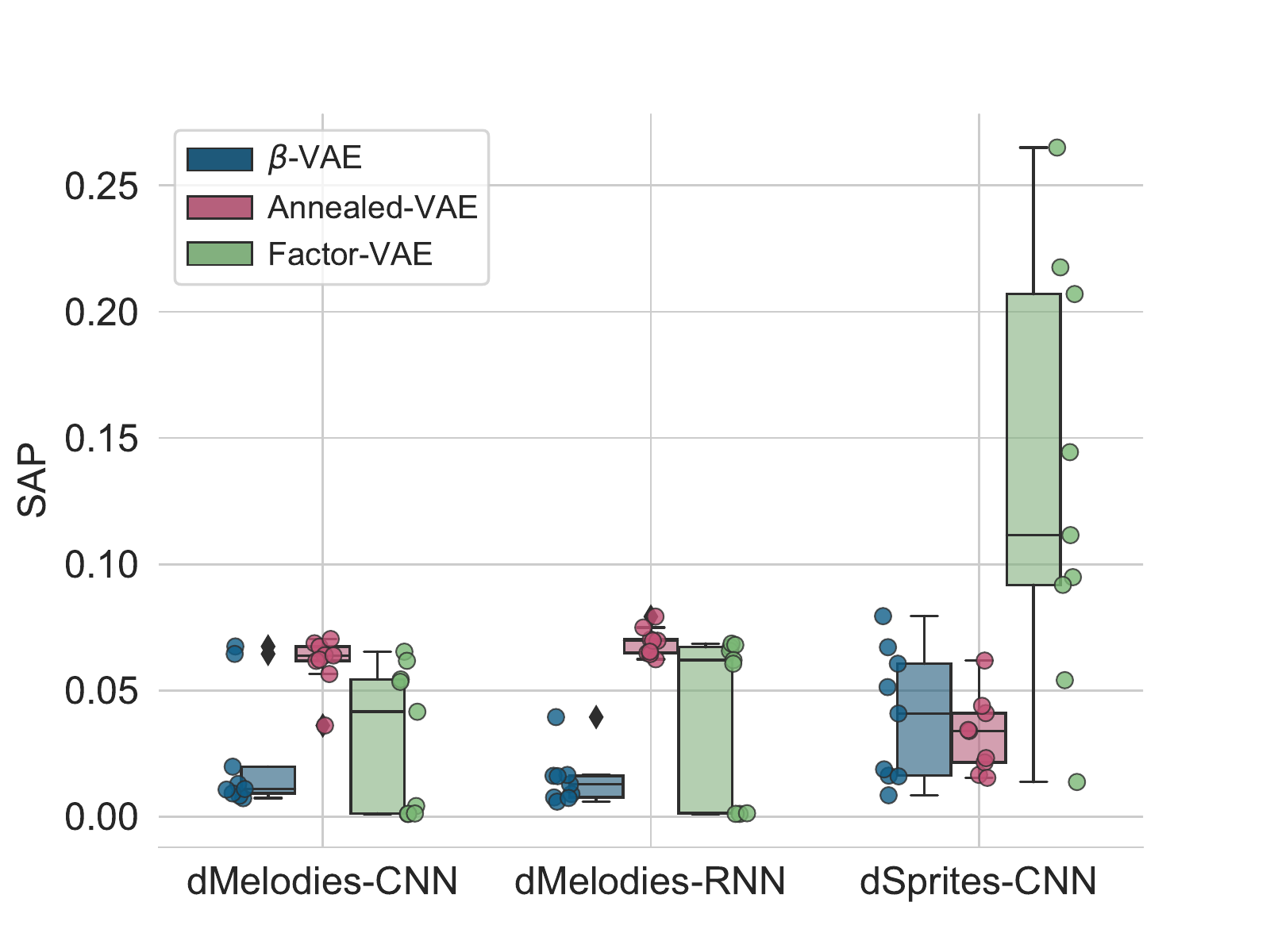} \\[\abovecaptionskip]
          \small (c) \textit{SAP Score}
        \end{tabular}
        \caption{Overall disentanglement performance (higher is better) of different methods on the dMelodies and dSprites datasets. Individual points denote results for different hyperparameter and random seed combinations. Please refer to supplementary material Sec.2.1 for the best hyperparameter settings.} 
        \label{fig:disent_results}
      \end{figure*}

      In this experiment, we present the comparative disentanglement performance of the different methods on dMelodies. The result for each method is aggregated across the different hyperparameters and random seeds. \figref{fig:disent_results} shows the results for all three disentanglement metrics. We group the trained models based on the architecture. The results for the dSprites dataset are also shown for comparison.  
      
      First, we compare the performance of different methods on dMelodies. Annealed-VAE shows better performance for MIG and SAP. These metrics indicate the ability of a method to ensure that each factor of variation is mapped to a single latent dimension. The performance in terms of Modularity is similar across the different methods. High Modularity indicates that each dimension of the latent space maps to only a single factor of variation. For dSprites, FactorVAE seems to be best method overall across metrics. However, the high variance in the results shows that choice of random seeds and hyperparameters is probably more important than the disentanglement method itself. This is in line with observations in previous studies \cite{locatello_challenging_2019}.

      Second, we observe no significant impact of model architecture on the disentanglement performance. For both the CNN and the hierarchical RNN-based VAE, the performance of all the different methods on dMelodies is comparable. This might be due to the relatively short sequence lengths used in dMelodies which do not fully utilize the capabilities of the hierarchical-RNN architecture (which has been shown to work well in learning long-term dependencies \cite{roberts_hierarchical_2018}). On the positive side, this indicates that the dMelodies dataset might be agnostic to the VAE-architecture.   

      Finally, we compare differences in the performance between the two datasets. In terms of MIG and SAP, the performance for dSprites is slightly better (especially for Factor-VAE), while for Modularity, performance across both datasets is comparable. However, once again, the differences are not significant. Looking at the disentanglement metrics alone, one might be tempted to conclude that the different methods are domain invariant. However, as the next experiments will show, there are significant differences. 

    \subsubsection{Reconstruction Fidelity} 
  
      \begin{figure}[t]
        \centering
        \includegraphics[width = 0.33\textwidth]{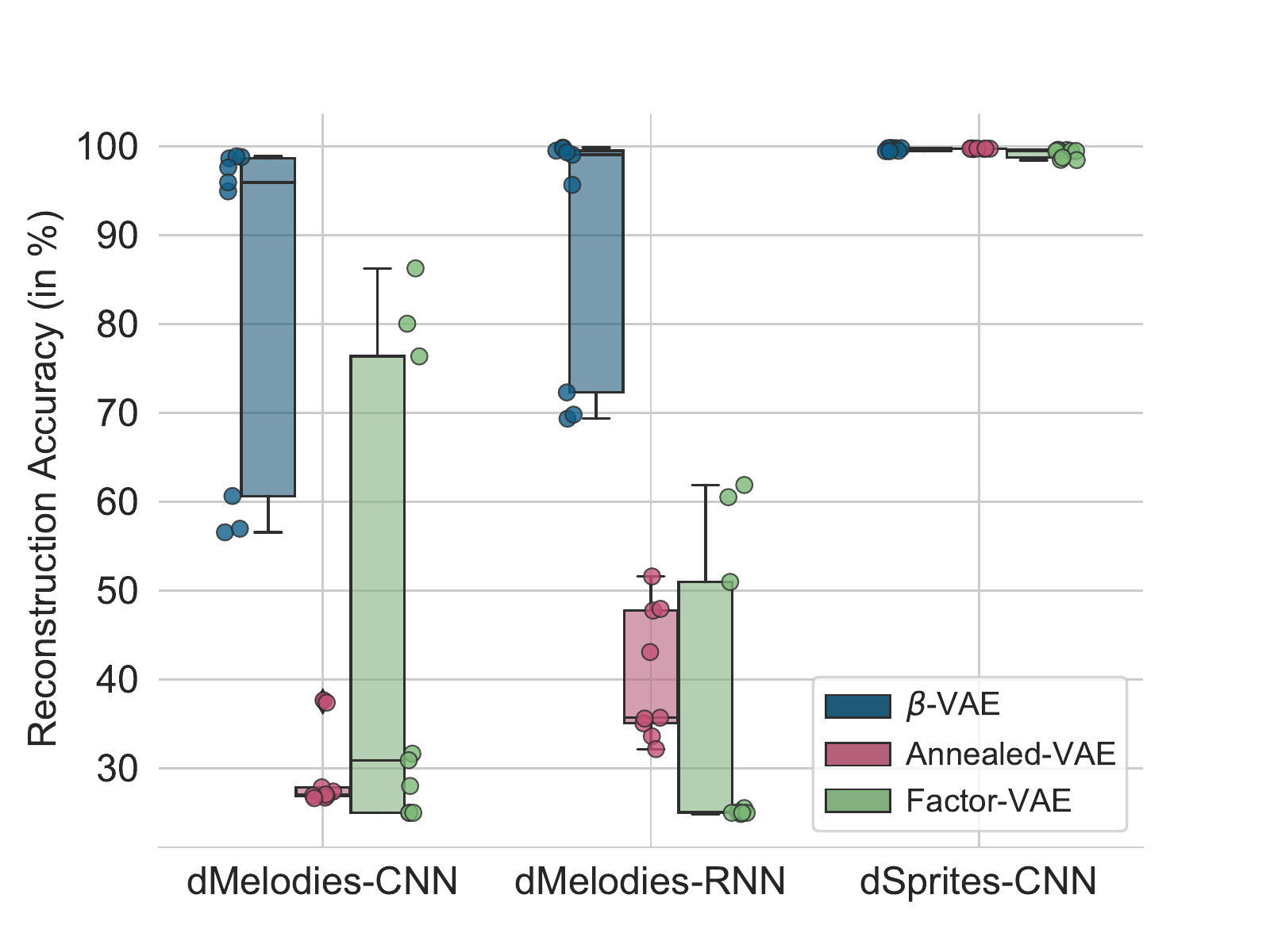}
        \caption{Overall reconstruction accuracies (higher is better) of the different methods on the dMelodies and dSprites datasets. Individual points denote results for different hyperparameter and random seed combinations.}
        \label{fig:recons_results}
      \end{figure}

      From a generative modeling standpoint, it is important that along with better disentanglement performance we also retain good reconstruction fidelity. This is measured using the reconstruction accuracy shown in \figref{fig:recons_results}. It is clear that all three methods fail to achieve a consistently good reconstruction accuracy on dMelodies. $\beta$-VAE gets an accuracy $\geq 90\%$ for some hyperparameter values (more on this in \secref{sec:hyper_param}). However, both Annealed-VAE and Factor-VAE struggle to cross a median-accuracy of $40\%$ (which would be unusable from a generative modeling perspective). The performance of the hierarchical RNN-based VAE is slightly better than the CNN-based architecture. In comparison, for dSprites, all three methods are able to consistently achieve better reconstruction accuracies. 

    \subsubsection{Sensitivity to Hyperparameters}
    \label{sec:hyper_param}
      
      \begin{figure*}[t]
        \centering
        \begin{tabular}{@{}c@{}}
          \includegraphics[width=0.33\textwidth]{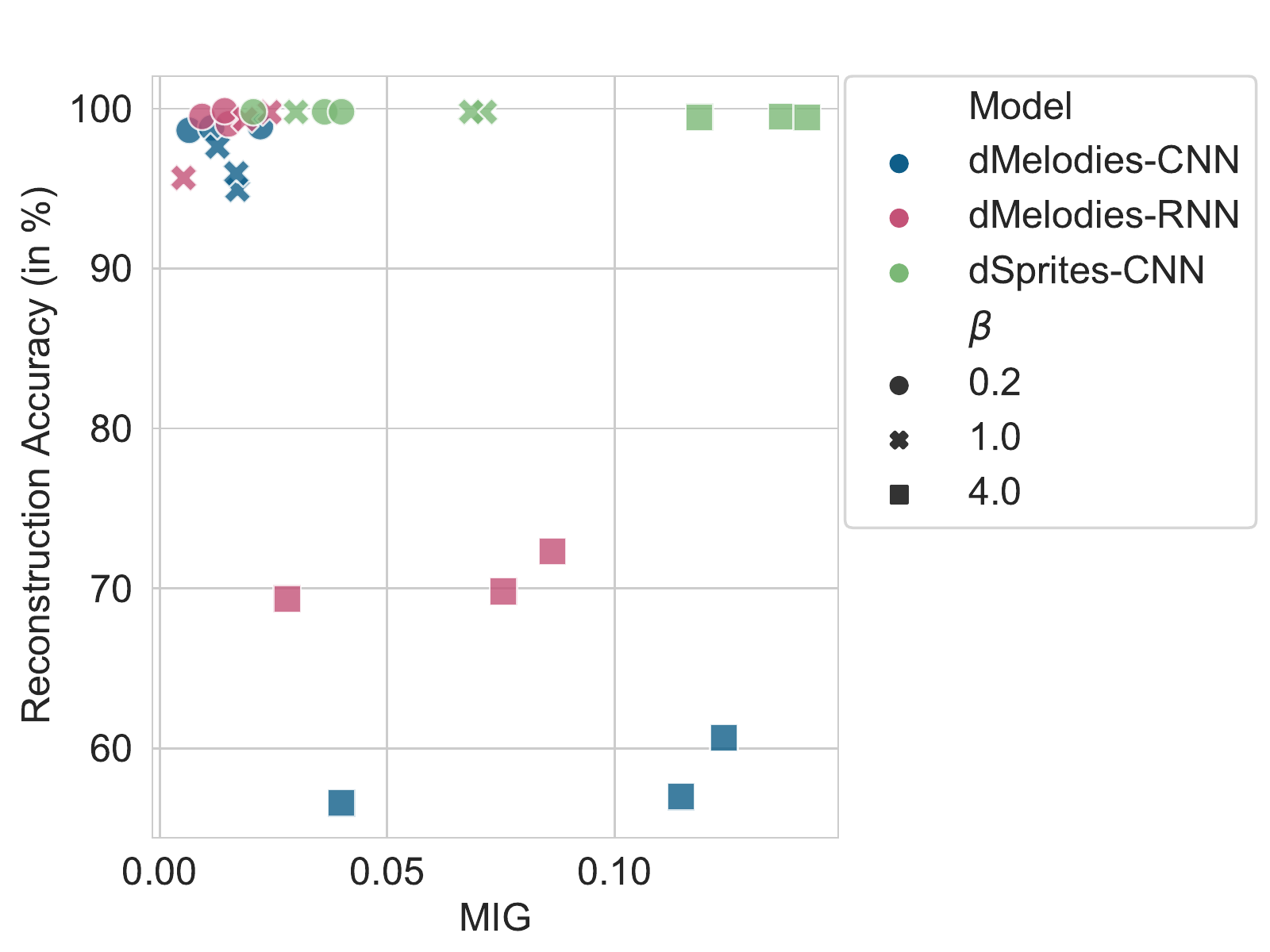} \\[\abovecaptionskip]
          \small (a) $\beta$-VAE: Varying $\beta$
        \end{tabular}
        \begin{tabular}{@{}c@{}}
          \includegraphics[width=0.33\textwidth]{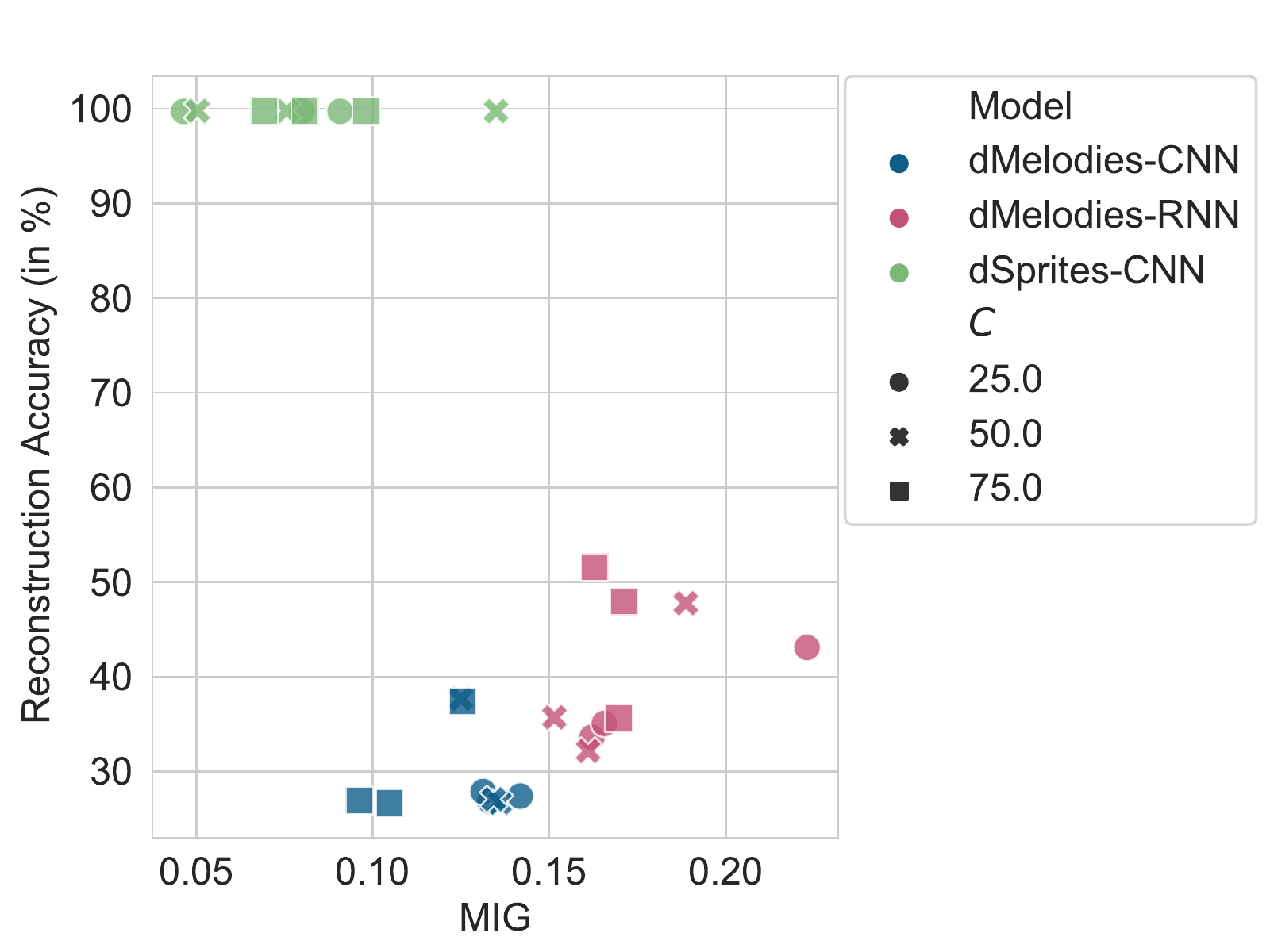} \\[\abovecaptionskip]
          \small (b) Annealed-VAE: Varying $C$
        \end{tabular}
        \begin{tabular}{@{}c@{}}
          \includegraphics[width=0.33\textwidth]{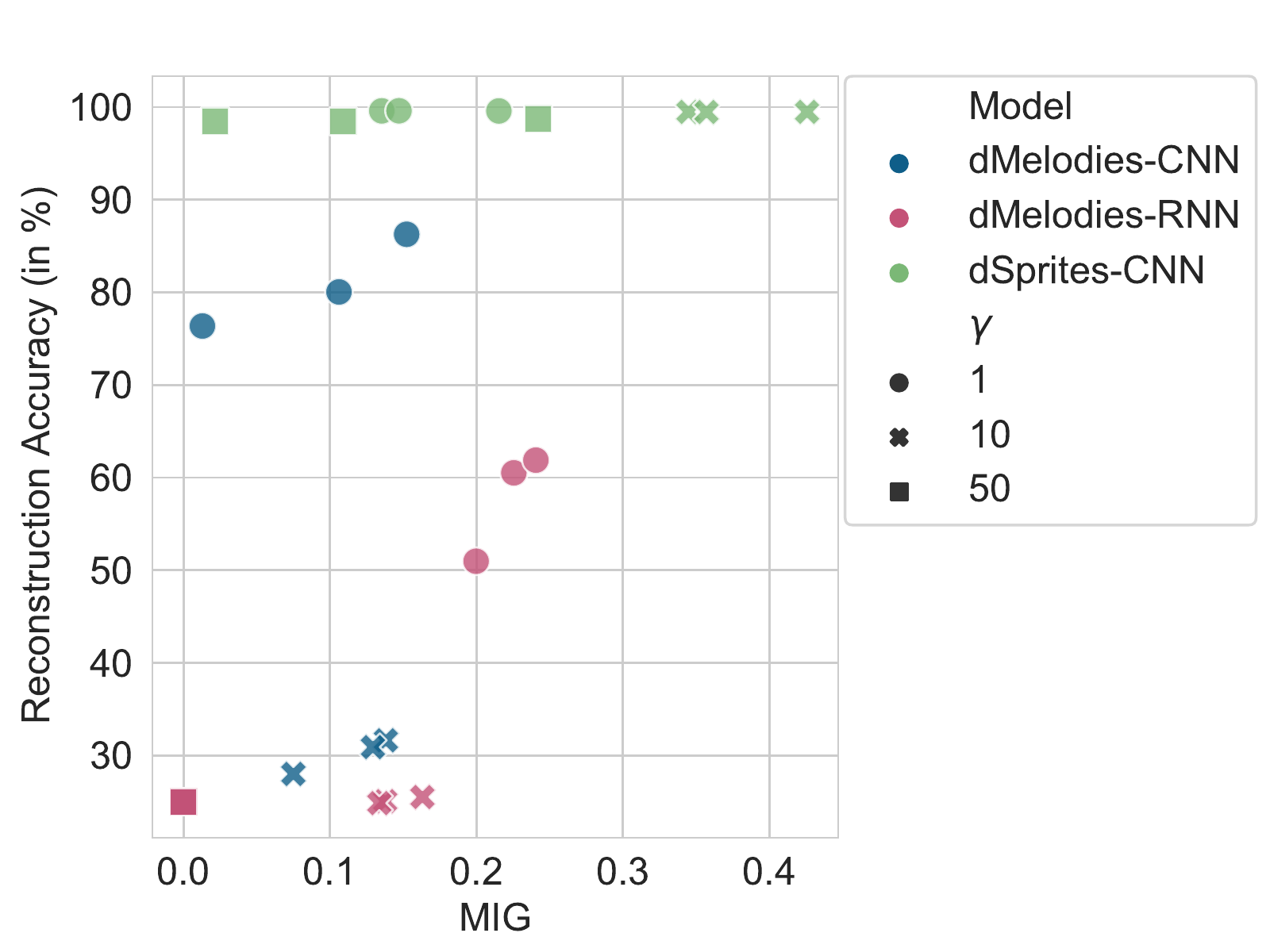} \\[\abovecaptionskip]
          \small (c) Factor-VAE: Varying $\gamma$
        \end{tabular}
        \caption{Effect of the hyperparameters on the different disentanglement methods. Overall, for improving disentanglement on dMelodies results in severe drop in reconstruction accuracy. The dSprites dataset does not suffer from this drawback.} 
        \label{fig:hyper_param_results}
      \end{figure*}
      
      The previous experiments presented aggregated results over the different hyperparameter values for each method. Next, we take a closer look at the individual impact of those hyperparameters, i.e., the effect of changing the hyperparameters on the disentanglement performance (MIG) and the reconstruction accuracy. \figref{fig:hyper_param_results} shows this in the form of scatter plots. The ideal models should lie on the top right corner of the plots (with high values of both reconstruction accuracy and MIG). 

      Models trained on dMelodies are very sensitive to hyperparameter adjustments. This is especially true for reconstruction accuracy. For instance, increasing $\beta$ for the $\beta$-VAE model improves MIG but severely reduces reconstruction performance. For Annealed-VAE and Factor-VAE there is a wider spread in the scatter plots. For Annealed-VAE, having a high capacity $C$ seems to marginally improve reconstruction (especially for the recurrent VAE). For FactorVAE, increasing $\gamma$ leads to a drop in both disentanglement and reconstruction. 

      Contrast this with the scatter plots for dSprites. For all three methods, the hyperparameters seem to only significantly affect the disentanglement performance. For instance, increasing $\beta$ and $\gamma$ (for $\beta$-VAE and FactorVAE, respectively) result in clear improvement in MIG. More importantly, however, there is no adverse impact on the reconstruction accuracy.

    \subsubsection{Factor-wise Disentanglement}

      \begin{figure}[t]
        \centering
        \includegraphics[width = 0.33\textwidth]{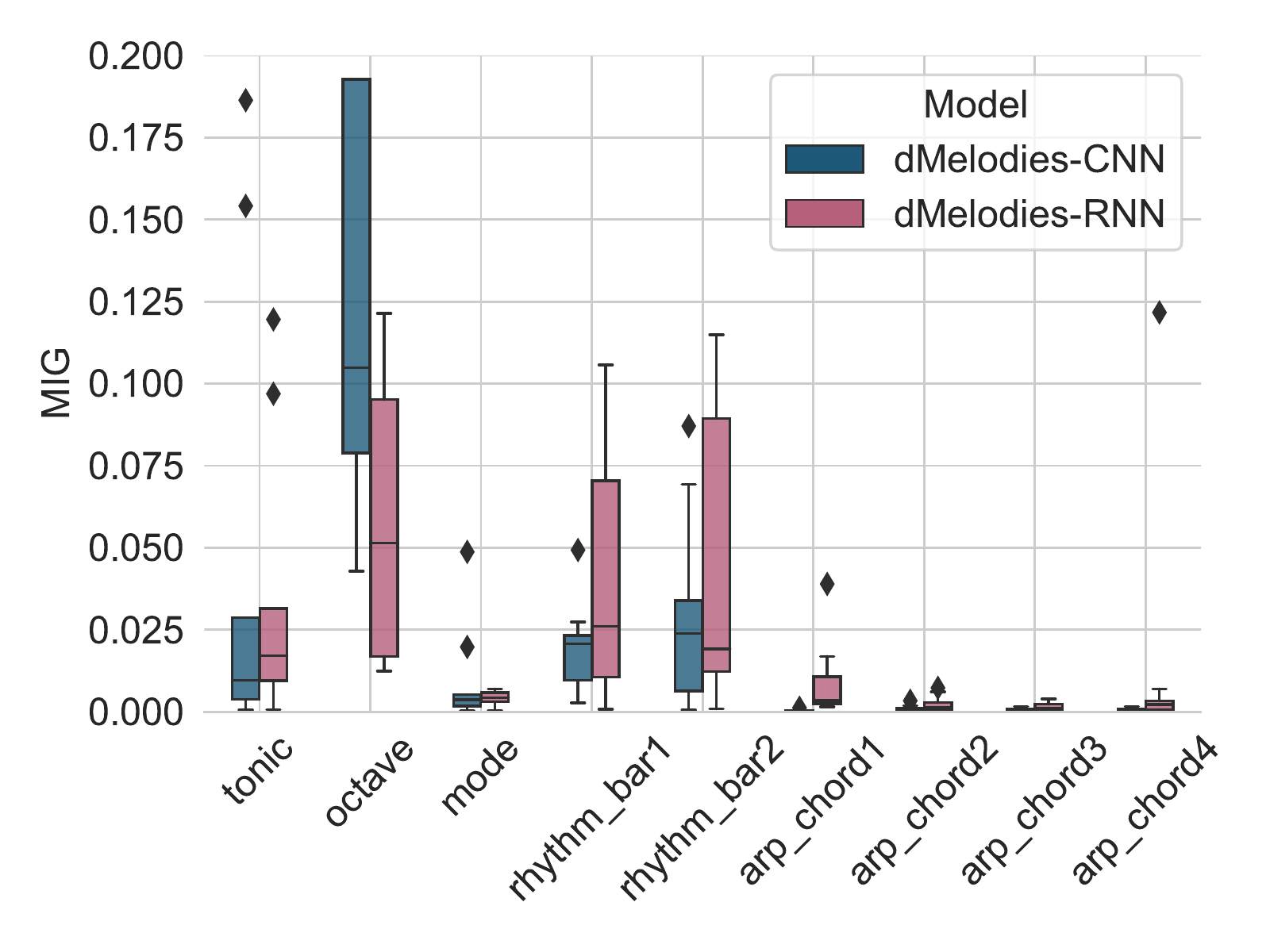}
        \caption{Factor-wise MIG for the $\beta$-VAE method.}
        \label{fig:factor_analysis}
      \end{figure}

      We also looked at how the individual factors of variation are disentangled. We consider the $\beta$-VAE model for this since it has the highest reconstruction accuracy. \figref{fig:factor_analysis} shows the factor-wise \textit{MIG} for both the CNN and RNN-based models. Factors corresponding to octave and rhythm are disentangled better. This is consistent with some recent research on disentangling rhythm \cite{yang2019deep, jiang2020transformer}. In contrast, the factors corresponding to the arpeggiation direction perform the worst. This might be due to their binary type. Similar analysis for the dSprites dataset reveals better disentanglement for the scale and position based factors. Additional results are provided in the supplementary material.

\section{Discussion}
\label{sec:results}
  As mentioned in \secref{sec:motivation}, disentanglement techniques have been shown to be sensitive to the choice of hyperparameters and random seeds \cite{locatello_challenging_2019}. The results obtained in our benchmarking experiments in the previous section using dMelodies seem to ascertain this even further. We find that methods which work well for image-based datasets do not extend directly to the music domain. When moving between domains, not only do we have to tune hyperparameters separately, but the model behavior may vary significantly when hyperparameters are changed. For instance, reconstruction fidelity is hardly effected by hyperparameter choice in the case of dSprites while for dMelodies it varies significantly. 
  While sensitivity to hyperparameters is expected in neural networks, this is also one of the main reasons for evaluating methods on more than one dataset, preferably from multiple domains. 
  
  \ashis{
    Some aspects of the dataset design, especially the nature of the factors of variation, might have affected our experimental results. While the factors of variation in dSprites are continuous (except the shape attribute), those for dMelodies span different data-types (categorical, ordinal and binary). This might make other types of models (such as VQ-VAEs \cite{oord2017vqvae}) more suitable. Another consideration is that some factors of variation (such as the arpeggiation direction and rhythm) effect only a part of the data. However, the effect of this on the disentanglement performance needs further investigation since we get good performance for rhythm but poor performance for arpeggiation direction.
    }

  Unsupervised methods for disentanglement learning have their own limitations and some degree of supervision might actually be essential \cite{locatello_challenging_2019}. It is still unclear if it is possible to develop general domain-invariant disentanglement methods. Consequently, supervised and semi-supervised methods have been garnering more attention \cite{pati19latent-reg,bouchacourt_multi-level_2018,hadjeres_glsr-vae_2017,Locatello2020Disentangling}. The dMelodies dataset can also be used to explore such methods for music-based tasks. There has been some work recently in disentangling musical attributes such as rhythm and melodic contours which are considered important from an interactive music generation perspective \cite{pati19latent-reg,akama_controlling_2019,yang2019deep}. Apart from the designed latent factors of variation, other low-level musical attributes such as rhythmic complexity and contours can also be computationally extracted using this dataset to meet task-specific requirements.
  
\section{Conclusion}
\label{sec:conclusion}
  This paper addresses the need for more diverse modes of data for studying disentangled representation learning by introducing a new music dataset for the task. The \textit{dMelodies} dataset comprises of more than 1 million data points of 2-bar melodies. The dataset is constructed based on fixed rules that maintain independence between different factors of variation, thus enabling researchers to use it for studying disentanglement learning. \ashis{Benchmarking experiments conducted using popular disentanglement learning methods show that existing methods do not achieve performance comparable to those obtained on an analogous image-based dataset. This showcases the need for further research on domain-invariant algorithms for disentanglement learning.}

\section{Acknowledgment}
  \ashis{The authors would like to thank Nvidia Corporation for their donation of a Titan V awarded as part of the GPU (Graphics Processing Unit) grant program which was used for running several experiments pertaining to this research.}  

\bibliography{2020-ISMIR-dMelodies}

\begin{thebibliography}{10}

\bibitem{akama_controlling_2019}
Taketo Akama.
\newblock Controlling {Symbolic} {Music} {Generation} {Based} {On} {Concept}
  {Learning} {From} {Domain} {Knowledge}.
\newblock In {\em Proc. of 20th International {Society} for {Music}
  {Information} {Retrieval} {Conference} ({ISMIR})}, Delft, The Netherlands,
  2019.

\bibitem{aubry_seeing_2014}
Mathieu Aubry, Daniel Maturana, Alexei~A. Efros, Bryan~C. Russell, and Josef
  Sivic.
\newblock Seeing {3D} {Chairs}: {Exemplar} {Part}-based {2D}-{3D} {Alignment}
  using a {Large} {Dataset} of {CAD} {Models}.
\newblock In {\em Proc. of {IEEE} {Conference} on {Computer} {Vision} and
  {Pattern} {Recognition} ({CVPR})}, pages 3762--3769, Columbus, Ohio, USA,
  2014.

\bibitem{bengio_representation_2013}
Yoshua Bengio, Aaron Courville, and Pascal Vincent.
\newblock Representation {Learning}: {A} {Review} and {New} {Perspectives}.
\newblock {\em IEEE Transactions on Pattern Analysis and Machine Intelligence},
  35(8), 2013.

\bibitem{bouchacourt_multi-level_2018}
Diane Bouchacourt, Ryota Tomioka, and Sebastian Nowozin.
\newblock Multi-{Level} {Variational} {Autoencoder}: {Learning} {Disentangled}
  {Representations} {From} {Grouped} {Observations}.
\newblock In {\em Proc. of 32nd {AAAI} {Conference} on {Artificial}
  {Intelligence}}, New Orleans, USA, 2018.

\bibitem{bretan15learning}
Mason Bretan and Larry Heck.
\newblock Learning semantic similarity in music via self-supervision.
\newblock In {\em Proc. of 20th International {Society} for {Music}
  {Information} {Retrieval} {Conference} ({ISMIR})}, Delft, The Netherlands,
  2019.

\bibitem{brunner_midi-vae_2018}
Gino Brunner, Andres Konrad, Yuyi Wang, and Roger Wattenhofer.
\newblock {MIDI}-{VAE}: {Modeling} {Dynamics} and {Instrumentation} of {Music}
  with {Applications} to {Style} {Transfer}.
\newblock In {\em Proc. of 19th International {Society} for {Music}
  {Information} {Retrieval} {Conference} ({ISMIR})}, Paris, France, 2018.

\bibitem{burges_3d-shapes_2020}
Chris Burgess and Kim Hyunjik.
\newblock 3d-shapes {Dataset}.
\newblock https://github.com/deepmind/3d-shapes, February 2020.
\newblock last accessed, 2nd April 2020.

\bibitem{burgess_understanding_2018}
Christopher~P. Burgess, Irina Higgins, Arka Pal, Loic Matthey, Nick Watters,
  Guillaume Desjardins, and Alexander Lerchner.
\newblock Understanding disentangling in $\beta$-{VAE}.
\newblock In {\em NIPS Workshop on Learning Disentangled Representations}, Long
  Beach, California, USA, 2017.

\bibitem{chen_isolating_2018}
Ricky T.~Q. Chen, Xuechen Li, Roger Grosse, and David Duvenaud.
\newblock Isolating {Sources} of {Disentanglement} in {Variational}
  {Autoencoders}.
\newblock In {\em Advances in {Neural} {Information} {Processing} {Systems} 31
  ({NeurIPS})}, Montréal, Canada, 2018.

\bibitem{chen_infogan_2016}
Xi~Chen, Yan Duan, Rein Houthooft, John Schulman, Ilya Sutskever, and Pieter
  Abbeel.
\newblock {InfoGAN}: {Interpretable} {Representation} {Learning} by
  {Information} {Maximizing} {Generative} {Adversarial} {Nets}.
\newblock In {\em Advances in {Neural} {Information} {Processing} {Systems} 29
  ({NeurIPS})}, pages 2172--2180, Barcelona, Spain, 2016.

\bibitem{choi2017transfer}
Keunwoo Choi, Gy{\"{o}}rgy Fazekas, Mark~B. Sandler, and Kyunghyun Cho.
\newblock Transfer learning for music classification and regression tasks.
\newblock In {\em Proc. of 18th International {Society} for {Music}
  {Information} {Retrieval} {Conference} ({ISMIR})}, pages 141--149, Suzhou,
  China, 2017.

\bibitem{connor2019representing}
Marissa Connor and Christopher Rozell.
\newblock Representing closed transformation paths in encoded network latent
  space.
\newblock In {\em Proc. of 34th AAAI Conference on Artificial Intelligence},
  New York, USA, 2020.

\bibitem{cuthbert_music21_2010}
Michael~Scott Cuthbert and Christopher Ariza.
\newblock music21: {A} {Toolkit} for {Computer}-{Aided} {Musicology} and
  {Symbolic} {Music} {Data}.
\newblock In {\em Proc. of 11th International {Society} for {Music}
  {Information} {Retrieval} {Conference} ({ISMIR})}, Utrecht, The Netherlands,
  2010.

\bibitem{donahue_semantically_2018}
Chris Donahue, Zachary~C. Lipton, Akshay Balsubramani, and Julian McAuley.
\newblock Semantically {Decomposing} the {Latent} {Spaces} of {Generative}
  {Adversarial} {Networks}.
\newblock In {\em Proc. of 6th International {Conference} on {Learning}
  {Representations} ({ICLR})}, Vancouver, Canada, 2018.

\bibitem{engel_latent_2017}
Jesse Engel, Matthew Hoffman, and Adam Roberts.
\newblock Latent {Constraints}: {Learning} to {Generate} {Conditionally} from
  {Unconditional} {Generative} {Models}.
\newblock In {\em Proc. of 5th International {Conference} on {Learning}
  {Representations} ({ICLR})}, Toulon, France, 2017.

\bibitem{gondal2019transfer}
Muhammad~Waleed Gondal, Manuel W{\"u}thrich, {\DJ}or{\dj}e Miladinovi{\'c},
  Francesco Locatello, Martin Breidt, Valentin Volchkov, Joel Akpo, Olivier
  Bachem, Bernhard Sch{\"o}lkopf, and Stefan Bauer.
\newblock On the transfer of inductive bias from simulation to the real world:
  a new disentanglement dataset.
\newblock In {\em Advances in {Neural} {Information} {Processing} {Systems} 32
  ({NeurIPS})}, pages 15740--15751, 2019.

\bibitem{gururani2019comparison}
Siddharth Gururani, Alexander Lerch, and Mason Bretan.
\newblock A comparison of music input domains for self-supervised feature
  learning.
\newblock In {\em Proc. of ICML Workshop on Machine Learning for Music
  Discovery Workshop (ML4MD), Extended Abstract}, Long Beach, California, USA,
  2019.

\bibitem{hadjeres2017deepbach}
Ga{\"e}tan Hadjeres, Fran{\c{c}}ois Pachet, and Frank Nielsen.
\newblock Deep{B}ach: {A} steerable model for {B}ach chorales generation.
\newblock In {\em Proc. of 34th International Conference on Machine Learning
  (ICML)}, pages 1362--1371, Sydney, Australia, 2017.

\bibitem{hadjeres_glsr-vae_2017}
Gaëtan Hadjeres, Frank Nielsen, and Francois Pachet.
\newblock {GLSR}-{VAE}: {Geodesic} latent space regularization for variational
  autoencoder architectures.
\newblock In {\em Proc. of {IEEE} {Symposium} {Series} on {Computational}
  {Intelligence} ({SSCI})}, pages 1--7, Hawaii, USA, 2017.

\bibitem{higgins_beta-vae_2017}
Irina Higgins, Loïc Matthey, Arka Pal, Christopher Burgess, Xavier Glorot,
  Matthew~M. Botvinick, Shakir Mohamed, and Alexander Lerchner.
\newblock $\beta$-{VAE}: {Learning} {Basic} {Visual} {Concepts} with a
  {Constrained} {Variational} {Framework}.
\newblock In {\em Proc. of 5th International {Conference} on {Learning}
  {Representations} ({ICLR})}, Toulon, France, 2017.

\bibitem{hsu2017unsupervised}
Wei-Ning Hsu, Yu~Zhang, and James Glass.
\newblock Unsupervised learning of disentangled and interpretable
  representations from sequential data.
\newblock In {\em Advances in Neural Information Processing Systems 30
  (NeurIPS)}, Long Beach, California, USA, 2017.

\bibitem{hsu2019disentangling}
Wei{-}Ning Hsu, Yu~Zhang, Ron~J. Weiss, Yu{-}An Chung, Yuxuan Wang, Yonghui Wu,
  and James~R. Glass.
\newblock Disentangling correlated speaker and noise for speech synthesis via
  data augmentation and adversarial factorization.
\newblock In {\em Proc. of {IEEE} International Conference on Acoustics, Speech
  and Signal Processing, {ICASSP} 2019}, Brighton, United Kingdom, 2019.

\bibitem{hung2019musical}
Yun-Ning Hung, I-Tung Chiang, Yi-An Chen, and Yi-Hsuan Yang.
\newblock Musical composition style transfer via disentangled timbre
  representations.
\newblock In {\em Proc. of 28th International Joint Conference on Artificial
  Intelligence (IJCAI)}, Macao, China, 2020.

\bibitem{jiang2020transformer}
Junyan Jiang, Gus~G Xia, Dave~B Carlton, Chris~N Anderson, and Ryan~H Miyakawa.
\newblock Transformer vae: A hierarchical model for structure-aware and
  interpretable music representation learning.
\newblock In {\em Proc. of IEEE International Conference on Acoustics, Speech
  and Signal Processing (ICASSP)}, pages 516--520, Barcelona, Spain, 2020.

\bibitem{kim_disentangling_2018}
Hyunjik Kim and Andriy Mnih.
\newblock Disentangling by {Factorising}.
\newblock In {\em Proc. of 35th International {Conference} on {Machine}
  {Learning} ({ICML})}, Stockholm, Sweeden, 2018.

\bibitem{kingma_adam_2015}
Diederik~P. Kingma and Jimmy Ba.
\newblock Adam: {A} {Method} for {Stochastic} {Optimization}.
\newblock In {\em Proc. of 3rd International {Conference} on {Learning}
  {Representations} ({ICLR})}, San Diego, USA, 2015.

\bibitem{kingma2014semi}
Diederik~P. Kingma, Danilo~J. Rezende, Shakir Mohamed, and Max Welling.
\newblock Semi-supervised learning with deep generative models.
\newblock In {\em Advances in Neural Information Processing Systems 27
  (NeurIPS)}, Montréal, Canada, 2014.

\bibitem{kingma_improved_2016}
Diederik~P. Kingma, Tim Salimans, Rafal Jozefowicz, Xi~Chen, Ilya Sutskever,
  and Max Welling.
\newblock Improved {Variational} {Inference} with {Inverse} {Autoregressive}
  {Flow}.
\newblock In {\em Advances in {Neural} {Information} {Processing} {Systems} 29
  ({NeurIPS})}, pages 4743--4751, Barcelona, Spain, 2016.

\bibitem{kingma_auto-encoding_2014}
Diederik~P. Kingma and Max Welling.
\newblock Auto-{Encoding} {Variational} {Bayes}.
\newblock In {\em Proc. of 2nd International {Conference} on {Learning}
  {Representations} ({ICLR})}, Banff, Canada, 2014.

\bibitem{kulkarni_deep_2015}
Tejas~D Kulkarni, William~F. Whitney, Pushmeet Kohli, and Josh Tenenbaum.
\newblock Deep {Convolutional} {Inverse} {Graphics} {Network}.
\newblock In {\em Advances in {Neural} {Information} {Processing} {Systems} 28
  ({NeurIPS})}, pages 2539--2547, Montréal, Canada, 2015.

\bibitem{kumar_variational_2017}
Abhishek Kumar, Prasanna Sattigeri, and Avinash Balakrishnan.
\newblock Variational {Inference} of {Disentangled} {Latent} {Concepts} from
  {Unlabeled} {Observations}.
\newblock In {\em Proc. of 5th International {Conference} of {Learning}
  {Representations} ({ICLR})}, Toulon, France, 2017.

\bibitem{lample_fader_2017}
Guillaume Lample, Neil Zeghidour, Nicolas Usunier, Antoine Bordes, Ludovic
  Denoyer, and Marc'Aurelio Ranzato.
\newblock Fader {Networks}:{Manipulating} {Images} by {Sliding} {Attributes}.
\newblock In {\em Advances in {Neural} {Information} {Processing} {Systems} 30
  ({NeurIPS})}, pages 5967--5976, Long Beach, California, USA, 2017.

\bibitem{lattner2019learning}
Stefan Lattner, Monika D{\"o}rfler, and Andreas Arzt.
\newblock Learning complex basis functions for invariant representations of
  audio.
\newblock In {\em Proc. of 20th International {Society} for {Music}
  {Information} {Retrieval} {Conference} ({ISMIR})}, Delft, The Netherlands,
  2019.

\bibitem{lee2020disentangled}
Jongpil Lee, Nicholas~J. Bryan, Justin Salamon, Zeyu Jin, and Juhan Nam.
\newblock Disentangled multidimensional metric learning for music similarity.
\newblock In {\em Proc. of IEEE International Conference on Acoustics, Speech
  and Signal Processing (ICASSP)}, pages 6--10, Barcelona, Spain, 2020.

\bibitem{liu_deep_2015}
Ziwei Liu, Ping Luo, Xiaogang Wang, and Xiaoou Tang.
\newblock Deep {Learning} {Face} {Attributes} in the {Wild}.
\newblock In {\em Proc. of {IEEE} {International} {Conference} on {Computer}
  {Vision} ({ICCV})}, pages 3730--3738, Santiago, Chile, 2015.

\bibitem{locatello_challenging_2019}
Francesco Locatello, Stefan Bauer, Mario Lucic, Gunnar Rätsch, Sylvain Gelly,
  Bernhard Schölkopf, and Olivier Bachem.
\newblock Challenging {Common} {Assumptions} in the {Unsupervised} {Learning}
  of {Disentangled} {Representations}.
\newblock In {\em Proc. of 36th International {Conference} on {Machine}
  {Learning} ({ICML})}, Long Beach, California, USA, 2019.

\bibitem{Locatello2020Disentangling}
Francesco Locatello, Michael Tschannen, Stefan Bauer, Gunnar Rätsch, Bernhard
  Schölkopf, and Olivier Bachem.
\newblock Disentangling factors of variations using few labels.
\newblock In {\em Proc. of 8th International Conference on Learning
  Representations (ICLR)}, Addis Ababa, Ethiopia, 2020.

\bibitem{luo2019learning}
Yin-Jyun Luo, Kat Agres, and Dorien Herremans.
\newblock Learning disentangled representations of timbre and pitch for musical
  instrument sounds using gaussian mixture variational autoencoders.
\newblock In {\em Proc. of 20th International {Society} for {Music}
  {Information} {Retrieval} {Conference} ({ISMIR})}, Delft, The Netherlands,
  2019.

\bibitem{matthey_dsprites_2017}
Loic Matthey, Irina Higgins, Demis Hassabis, and Alexander Lerchner.
\newblock {dSprites}: {Disentanglement} testing {Sprites} dataset.
\newblock https://github.com/deepmind/dsprites-dataset, 2017.
\newblock last accessed, 2nd April 2020.

\bibitem{pati19latent-reg}
Ashis Pati and Alexander Lerch.
\newblock Latent space regularization for explicit control of musical
  attributes.
\newblock In {\em Proc. of ICML Workshop on Machine Learning for Music
  Discovery Workshop (ML4MD), Extended Abstract}, Long Beach, California, USA,
  2019.

\bibitem{pati_learning_2019}
Ashis Pati, Alexander Lerch, and Gaëtan Hadjeres.
\newblock Learning to {Traverse} {Latent} {Spaces} for {Musical} {Score}
  {Inpainting}.
\newblock In {\em Proc. of 20th International {Society} for {Music}
  {Information} {Retrieval} {Conference} ({ISMIR})}, Delft, The Netherlands,
  2019.

\bibitem{reed_deep_2015}
Scott~E Reed, Yi~Zhang, Yuting Zhang, and Honglak Lee.
\newblock Deep {Visual} {Analogy}-{Making}.
\newblock In {\em Advances in {Neural} {Information} {Processing} {Systems} 28
  ({NeurIPS})}, pages 1252--1260, Montréal, Canada, 2015.

\bibitem{ridgeway_learning_2018}
Karl Ridgeway and Michael~C Mozer.
\newblock Learning {Deep} {Disentangled} {Embeddings} {With} the
  {F}-{Statistic} {Loss}.
\newblock In {\em Advances in {Neural} {Information} {Processing} {Systems} 31
  ({NeurIPS})}, pages 185--194, Montréal, Canada, 2018.

\bibitem{roberts_hierarchical_2018}
Adam Roberts, Jesse Engel, Colin Raffel, Curtis Hawthorne, and Douglas Eck.
\newblock A {Hierarchical} {Latent} {Vector} {Model} for {Learning}
  {Long}-{Term} {Structure} in {Music}.
\newblock In {\em Proc. of 35th International {Conference} on {Machine}
  {Learning} ({ICML})}, Stockholm, Sweeden, 2018.

\bibitem{rubenstein_learning_2018}
Paul Rubenstein, Bernhard Scholkopf, and Ilya Tolstikhin.
\newblock Learning {Disentangled} {Representations} with {Wasserstein}
  {Auto}-{Encoders}.
\newblock In {\em Proc. of 6th International {Conference} on {Learning}
  {Representations} ({ICLR}), {Workshop} {Track}}, Vancouver, Canada, 2018.

\bibitem{siddharth2017learning}
N.~Siddharth, Brooks Paige, Jan-Willem van~de Meent, Alban Desmaison, Noah~D.
  Goodman, Pushmeet Kohli, Frank Wood, and Philip~H.S. Torr.
\newblock Learning disentangled representations with semi-supervised deep
  generative models.
\newblock In {\em Advances in Neural Information Processing Systems 30
  (NeurIPS)}, Long Beach, California, USA, 2017.

\bibitem{toussaint_mathematical_2002}
Godfried Toussaint.
\newblock A {Mathematical} {Analysis} of {African}, {Brazilian} and {Cuban}
  {Clave} {Rhythms}.
\newblock In {\em Proc. of {BRIDGES}: {Mathematical} {Connections} in {Art},
  {Music} and {Science}}, pages 157--168, 2002.

\bibitem{oord2017vqvae}
Aaron van~den Oord, Oriol Vinyals, and koray kavukcuoglu.
\newblock Neural discrete representation learning.
\newblock In {\em Advances in Neural Information Processing Systems 30
  (NeurIPS)}, pages 6306--6315. Long Beach, California, USA, 2017.

\bibitem{yang2019deep}
Ruihan Yang, Dingsu Wang, Ziyu Wang, Tianyao Chen, Junyan Jiang, and Gus Xia.
\newblock Deep music analogy via latent representation disentanglement.
\newblock In {\em Proc. of 20th International {Society} for {Music}
  {Information} {Retrieval} {Conference} ({ISMIR})}, Delft, The Netherlands,
  2019.

\bibitem{kexin2018neural}
Kexin Yi, Jiajun Wu, Chuang Gan, Antonio Torralba, Pushmeet Kohli, and Josh
  Tenenbaum.
\newblock Neural-symbolic vqa: Disentangling reasoning from vision and language
  understanding.
\newblock In {\em Advances in Neural Information Processing Systems 31
  {(NeurIPS)}}. Montréal, Canada, 2018.

\end{thebibliography}

\end{document}